\documentclass[10pt,twocolumn,letterpaper]{article}

\usepackage{iccv}
\usepackage{times}
\usepackage{epsfig}
\usepackage{graphicx}
\usepackage{amsmath}
\usepackage{amssymb}

\usepackage{subcaption}
\usepackage{multirow} 
\usepackage{makecell}
\usepackage{booktabs}

\usepackage{algorithm}
\usepackage{algpseudocode}
\algtext*{EndFor}
\algtext*{EndIf}

\usepackage[breaklinks=true,bookmarks=false,hidelinks]{hyperref}

\graphicspath{ {images/} {images-suppl/} }

\iccvfinalcopy 

\def\iccvPaperID{2560} 

\ificcvfinal\fi

\RequirePackage{tcolorbox}

\newcommand\blfootnote[1]{%
  \begingroup
  \renewcommand\thefootnote{}\footnote{#1}%
  \addtocounter{footnote}{-1}%
  \endgroup
}

\begin{document}

\title{MAS: Towards Resource-Efficient Federated Multiple-Task Learning}

\author{Weiming Zhuang$^{1*}$ \quad Yonggang Wen$^{2}$  \quad Lingjuan Lyu$^{1}$ \quad Shuai Zhang$^{3}$\\
$^{1}$Sony AI, $^{2}$Nanyang Technological University, 
$^{3}$SenseTime Research \\
{\tt\small \{weiming.zhuang,lingjuan.lv\}@sony.com,ygwen@ntu.edu.sg,zhangshuai@sensetime.com}
}

\maketitle
\ificcvfinal\thispagestyle{empty}\fi

\begin{abstract}
Federated learning (FL) is an emerging distributed machine learning method that empowers in-situ model training on decentralized edge devices. However, multiple simultaneous FL tasks could overload resource-constrained devices. In this work, we propose the first FL system to effectively coordinate and train multiple simultaneous FL tasks. We first formalize the problem of training simultaneous FL tasks. Then, we present our new approach, MAS (Merge and Split), to optimize the performance of training multiple simultaneous FL tasks. MAS starts by merging FL tasks into an all-in-one FL task with a multi-task architecture. After training for a few rounds, MAS splits the all-in-one FL task into two or more FL tasks by using the affinities among tasks measured during the all-in-one training. It then continues training each split of FL tasks based on model parameters from the all-in-one training. Extensive experiments demonstrate that MAS outperforms other methods while reducing training time by 2$\times$ and reducing energy consumption by 40\%. We hope this work will inspire the community to further study and optimize training simultaneous FL tasks.
\end{abstract}


\section{Introduction}
Federated learning (FL) \cite{fedavg} has attracted considerable attention as it enables privacy-preserving distributed model training among decentralized devices. It is empowering growing numbers of applications in both academia and industry, such as medical imaging analysis \cite{li2019brain-tumor1,sheller2018brain-tumor2}, Google Keyboard \cite{hard2018gboard}, and autonomous vehicles \cite{zhang2021auto,posner2021vehicular}. Among them, some applications contain multiple application tasks. For example, autonomous vehicles are related to multiple resource-intensive computer vision (CV) tasks, including lane detection, object detection, and segmentation \cite{janai2020cvforauto}. \blfootnote{*most of the work done in S-Lab, Nanyang Technological University}

In fact, the majority of edge devices (e.g., NVIDIA Jeston TX2 and AGX Xavier) can only support one FL task at a time \cite{liu2019performance}. Multiple simultaneous FL tasks on the same device could overwhelm its memory, computation, and power capacities. Thus, it is important to navigate solutions to well coordinate these simultaneous FL tasks.

\begin{figure}[t!]
  \centering
  \includegraphics[width=0.45\textwidth]{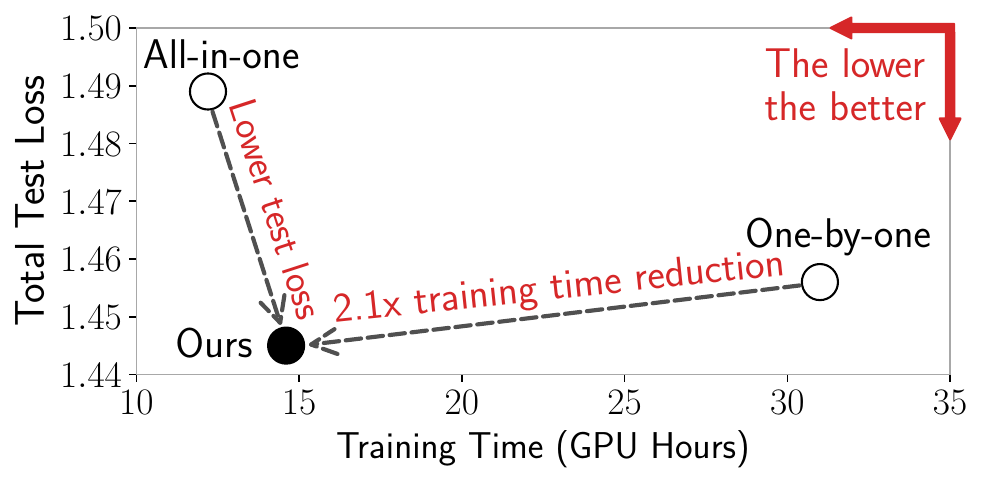}
 \caption{Existing methods suffer from a trade-off between training time and test loss (lower test loss means better performance) when training 9 simultaneous FL tasks, whereas our method navigates a sweet point, achieving the best test loss with 2.1$\times$ training time reduction. One-by-one trains FL tasks one after another; All-in-one combines tasks into a multi-task learning network before training it in FL.}
 \label{fig:motivation}
\end{figure}

A plethora of research on FL are mainly devoted to addressing challenges such as statistical heterogeneity \cite{fedprox, wang2020fedma}, system heterogeneity \cite{chai2020tifl, yang2021characterizing, zhang2022dense}, communication efficiency \cite{karimireddy2020scaffold, zhu2021delayed, li2022mocosfl, wu2022communication}, and privacy issues \cite{bagdasaryan2020backdoor, huang2021evaluating}. 
Most of the existing works only focus on one FL task, 
overlooking the fact that certain applications, such as self-driving cars or intelligent manufacturing robots, need to tackle multiple FL tasks simultaneously \cite{janai2020cvforauto,FLAIROP}.
To address this issue, Bonawitz et.al~\cite{bonawitz2019flsys} designed multi-tenancy to prevent simultaneous FL tasks from overloading devices. However, their proposed method is slow in training because it regards these tasks as independent training tasks and trains them sequentially. This \textit{one-by-one} training method only considers the differences among tasks, neglecting potential synergies.

Another intuitive solution is to adopt multi-task learning (MTL) to train multiple FL tasks by combining these tasks into an \textit{all-in-one} neural network. This network has one encoder shared among tasks and multiple task-specific decoders. It could prevent overloading devices and speeds up the training process as only one neural network is trained. However, it could result in worse performance because not all tasks are beneficial to the others when training together \cite{kang2011learning,zhao2018modulation}. Simply combining FL tasks together only takes into account their synergies while overlooking their distinctions. Figure \ref{fig:motivation} shows that either the all-in-one (only considers task synergies) or the one-by-one method (only considers task differences) suffers from a trade-off between training time and test loss.

In this work, we propose MAS (i.e. \underline{M}erge \underline{a}nd \underline{S}plit), the first FL system to effectively coordinate and train multiple simultaneous FL tasks under resource constraints by considering both synergies and differences among these tasks. We first formalize the problem of training multiple simultaneous FL tasks. To address this problem, we introduce MAS to optimize the performance. Specifically, MAS starts by merging these FL tasks into an all-in-one FL task with a multi-task architecture, which shares common layers and has specialized layers for each task. After training the all-in-one FL task for certain rounds, MAS splits this all-in-one task into two or more FL tasks based on their synergies and differences measured by affinity scores during training. Lastly, MAS continues training each split of FL tasks with models trained in the all-in-one process. 

Figure \ref{fig:motivation} shows that MAS achieves the best test loss with 2.1$\times$ training time reduction compared to the one-by-one method on training nine FL tasks. We also demonstrate that it reduces energy consumption by over 40\% while achieving superior performance to other methods via extensive experiments on three different sets of FL tasks. We believe that MAS is beneficial for many real-world applications such as autonomous vehicles and robotics. We summarize our contributions as follows:

\begin{itemize}
  \item We formalize the problem of training multiple simultaneous FL tasks. To the best of our knowledge, we are the first to conduct an in-depth investigation into the training of multiple simultaneous FL tasks.
  \item We propose MAS, a new FL system to effectively coordinate and train simultaneous FL tasks by considering both synergies and differences among these tasks.
  \item We establish baselines for training multiple simultaneous FL tasks and demonstrate that MAS elevates performance with significantly less training time and energy consumption via extensive empirical studies.
\end{itemize}

\section{Related Work}
\label{sec:related-work}

In this section, we provide a literature review of federated learning and multi-task learning.

\textbf{Federated Learning} emerges as a privacy-aware and distributed learning paradigm that uses a central server to coordinate multiple decentralized clients to train models \cite{fedavg,kairouz2021advances}. The majority of studies aim to address the challenges of FL, including statistical heterogeneity \cite{fedprox, wang2020fedma, wang2020fednova, zhuang2020fedreid, deng2021adaptive, zhang2021parameterized,zhuang2022fedema, tan2023taming, zhang2023addressing, fan2022fedskip, zhuang2023normalization}, system heterogeneity \cite{chai2020tifl, yang2021characterizing, liu2022no}, communication efficiency \cite{fedavg, jakub2016communication, karimireddy2020scaffold, zhu2021delayed}, and privacy concerns \cite{bagdasaryan2020backdoor, huang2021evaluating}. Numerous methods are proposed to cluster FL clients into groups to address statistical heterogeneity \cite{ghosh2020efficient, ouyang2021clusterfl, zhuang2022fedreid}. They aim to cluster models that are trained on clients with similar distribution, whereas our proposed MAS differs fundamentally from these methods as it splits simultaneous FL tasks into groups. Several other attempts have been made \cite{smith2017fedmultil,marfoq2021federated} on \textit{federated multi-task learning} in order to learn personalized models to tackle statistical heterogeneity. These personalized FL methods mainly focus on training one FL task of an application in a client. Training multiple simultaneous FL tasks is rarely explored. 
The prior work \cite{bonawitz2019flsys} designs multi-tenancy in an FL system to schedule and train these tasks sequentially. This one-by-one method is slow in training and only considers the differences among these FL tasks. 

\textbf{Multi-task Learning} is a popular machine learning approach to learn models that can generalize on multiple tasks \cite{thrun1995learning,zhang2021survey}. A plethora of studies investigate parameter sharing approaches that share common layers of a similar architecture \cite{caruana1997multitask,eigen2015predicting,bilen2016integrated,nekrasov2019real}. Besides, many studies employ new techniques to address the negative transfer problem \cite{kang2011learning,zhao2018modulation} among tasks, including soft parameter sharing \cite{duong2015low,misra2016cross}, neural architecture search \cite{lu2017fully,huang2018gnas,vandenhende2019branched,guo2020learning,sun2020adashare}, and dynamic loss reweighting strategies \cite{kendall2018multi,chen2018gradnorm,yu2020gradient}. Instead of training all tasks together, task grouping trains only similar tasks together. The early works of task grouping \cite{kang2011learning,kumar2012learning} are not adaptable to DNN. Recently, several studies analyze task similarity \cite{standley2020which} and task affinities \cite{fifty2021tag} for task grouping. 
The state-of-the-art task grouping methods \cite{standley2020which,fifty2021tag}, however, are unsuitable for training multiple simultaneous FL tasks because they mainly focus on inference efficiency. They would train a task multiple times as their task groups always contain overlapped tasks.
This motivates us to exploit task merging and task splitting to group and train multiple simultaneous FL tasks.

\section{Method}

In this section, we start by providing problem definition for training multiple simultaneous FL tasks. Then, we propose Merge and Split (MAS) method that first merges tasks into an all-in-one FL task and then splits it into two or more splits for further training.

\subsection{Problem Definition}
\label{sec:problem-setup}

In the federated learning setting, the majority of studies consider optimizing the following problem:

\begin{equation}
  \min_{\omega \in \mathbb{R}^d} f(\omega) := \sum_{k=1}^K p_k f_k(\omega) := \sum_{k=1}^K p_k \mathbb{E}_{\xi_k \sim \mathcal{D}_k}[f_k(\omega;\xi_k)],
  \label{eq:fl}
\end{equation}
where $\omega$ is the optimization variable, $K$ is the number of selected clients to execute training, $f_k(\omega)$ is the loss function of client $k$, $p_k$ is the weight of client $k$ in model aggregation, and $\xi_k$ is the training data sampled from data distribution $\mathcal{D}_k$ of client $k$. \verb+FedAvg+ \cite{fedavg} is a popular federated learning algorithm, which sets $p_k$ to be proportional to the dataset size of client $k$.

In fact, Equation \ref{eq:fl} only illustrates the objective of training a single FL task. In real-world scenarios, an FL server could receive multiple simultaneous FL tasks, denoted as a set $\mathcal{A} = \{\alpha_1,\alpha_2,\dots,\alpha_n\}$. These tasks aim to train a set of models $\mathcal{W} = \{\omega_1,\omega_2,\dots,\omega_n\}$, where each model $\omega_i$ is for task $\alpha_i$. By defining $\mathcal{M}(\alpha_i;\omega_i)$ as the performance measurement of each FL task $\alpha_i$, the overall objective is to maximize the performance of all FL tasks $\sum_{i=1}^n \mathcal{M}(\alpha_i;\omega_i)$ with minimum training time, under the constraint that each client $k$ has limited memory budget and computation budget. These budgets constrain the number of simultaneous FL tasks $n_k$ on client $k$. Besides, as devices have limited battery life, it is important to minimize the energy consumption and training time to obtain $\mathcal{W}$ for FL tasks $\mathcal{A}$. 

In this work, we assume that each client can execute one FL task at a time ($n_k$ = 1). This is common for the majority of current edge devices\footnote{Edges devices, e.g., NVIDIA Jetson TX2 and AGX Xavier, have only one GPU; GPU virtualization \cite{hong2017gpu} that enables concurrent training on the same GPU currently are mainly for the cloud stack.}. Besides, we assume that the models $\mathcal{W} = \{\omega_1,\omega_2,\dots,\omega_n\}$ share the same backbone architecture but have different decoder architectures. This is a practical assumption for many real-world applications and industrial practices \cite{apple-mtl, FLAIROP}.

\begin{figure}[t!]
  \centering
  \includegraphics[width=0.45\textwidth]{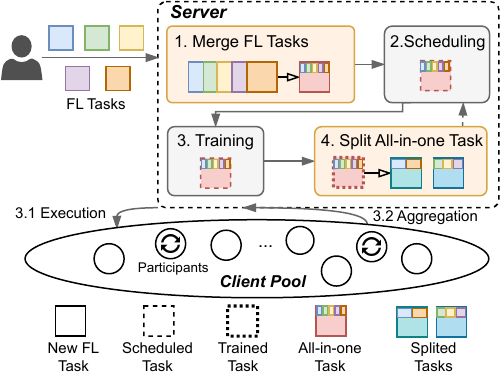}
 \caption{The architecture and workflow of our proposed \underline{M}erge \underline{a}nd \underline{S}plit (MAS). The server first merges multiple simultaneous FL tasks into an all-in-one FL task. After scheduling and training this task in FL for certain rounds, MAS splits the all-in-one FL task into two or more splits based on task affinities measured during training. It considers both synergies and differences among FL tasks by splitting the tasks with higher synergies in the same split.}
 \label{fig:mas}
\end{figure}

\begin{figure}[t!]
  \centering
  \includegraphics[width=0.45\textwidth]{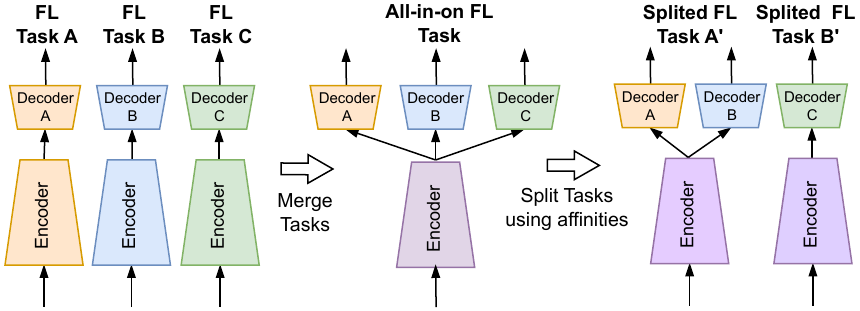}
 \caption{Illustration of network architecture changes in MAS. Initially, each FL task employs an encoder and a decoder. MAS first merges these FL tasks into an all-in-one FL task with a multi-task architecture. Then, it splits the all-in-one FL task into two or more splits.}
 \label{fig:architecture}
\end{figure}

\subsection{Architecture Overview}

Figure \ref{fig:mas} depicts the architecture overview and training process of our proposed MAS, which trains  simultaneous FL tasks efficiently by considering both synergies and differences among these tasks. It contains a server to coordinate FL tasks and a pool of clients to execute training. 

The training process of MAS is as follows: 1) The server receives multiple FL tasks $\mathcal{A} = \{\alpha_1,\alpha_2,\dots,\alpha_n\}$ to train models $\mathcal{W} = \{\omega_1,\omega_2,\dots,\omega_n\}$ and merges these FL tasks into an all-in-one FL task $\alpha_0$ with a multi-task model $\phi$. 2) The server schedules $\alpha_0$ to train $\phi$. 3) The server iteratively selects $K$ clients from the client pool to train $\alpha_0$ through FL process for $R_0$ rounds. In each round, the server sends model $\phi$ to the selected clients; the clients train $\phi$ and calculate task affinity scores $\hat{\mathcal{S}}$ before uploading these training updates to the server. 4) The server uses the affinity scores $\hat{\mathcal{S}}$ to split the all-in-one FL task $\alpha_0$ into two or more FL task splits $\{\mathcal{A}_1,\mathcal{A}_2,\dots\}$, where each split trains non-overlapping subset of $\mathcal{W}$. The number of splits can be determined by the inference budget for the number of concurrent models. 5) The server iterates steps 2 and 3 to train $\mathcal{A}_j$. We summarize MAS in Algorithm \ref{algo:mufl} and the network architecture changes in Figure \ref{fig:architecture}. 

\subsection{Merge FL Tasks Into an All-in-one Task}

The server receives multiple FL tasks and merges them into an all-in-one FL task with multi-task architecture. This is based on the practical assumption discussed in Section \ref{sec:problem-setup} that the models of these FL tasks could share similar model architecture -- sharing the same backbone architecture and having task-specific decoders. We can merge these FL tasks into an all-in-one FL task $\alpha_0$ that trains a multi-task model $\phi = \{\theta_s\} \cup \{\theta_{\alpha_i} | \alpha_i \in \mathcal{A} \}$, where $\theta_s$ is the shared model parameters and $\theta_{\alpha_i}$ is the specific parameters for FL task $\alpha_i \in \mathcal{A}$. The loss function for training the all-in-one FL task in each client is as followed: 
\begin{equation}
  \mathcal{L}(\mathcal{X}, \theta_s, \{\theta_{\alpha_i}\}) = \sum_{\alpha_i \in \mathcal{A}} \mathcal{L}_{\alpha_i}(\mathcal{X}, \theta_s, \theta_{\alpha_i}),
\end{equation}
where $\mathcal{X}$ is batch of data and $\mathcal{L}_{\alpha_i}$ denotes the loss function of each FL task $\alpha_i \in \mathcal{A}$. This formulation is generally applicable to different loss weights, but we set the loss weight to one for simplicity of notations.

Merging FL tasks into an all-in-one task effectively reduces the training time (Figure \ref{fig:motivation}), as we only need to train one task instead of multiple tasks sequentially. However, simply training with the all-in-one FL task leads to unsatisfactory performance on total test loss, because it only considers synergies among tasks, neglecting the negative transfer problems \cite{kang2011learning,zhao2018modulation} in multi-task learning. Consequently, we further propose to split the all-in-one FL task considering both synergies and differences among these tasks.


\subsection{Split All-in-one FL Task Into Multiple Splits}

MAS divides the all-in-one FL task $\alpha_0$ into multiple splits after it is trained for certain rounds. Essentially, we aim to split $\mathcal{A} = \{\alpha_1,\alpha_2,\dots,\alpha_n\}$ into multiple non-overlapping groups such that FL tasks within a group have better synergy. Let $\{\mathcal{A}_1, \mathcal{A}_2,\dots,\mathcal{A}_m\}$ be subsets of $\mathcal{A}$, we aim to find a disjoint set $I$ of $\mathcal{A}$, where $I \subseteq \{1, 2, \dots, m\}$, $|I| \leq |\mathcal{A}|$, $\bigcup_{j\in I}\mathcal{A}_j = \mathcal{A}$, and $\bigcap_{j\in I}\mathcal{A}_j = \emptyset$. Each split $\mathcal{A}_j$ trains a model $\phi_j = \{\theta_{s}^{j}\} \cup \{\theta_{\alpha_i} | \alpha_i \in \mathcal{A}_j\}$, which is a multi-task network when $\mathcal{A}_j$ contains more than one FL task, where $\theta_s^j$ is the shared model parameters and $\theta_{\alpha_i}$ is the specific parameters for FL task $\alpha_i \in \mathcal{A}_j$. The core question is how to determine set $I$ to split these FL tasks considering their synergies and differences.

Inspired by TAG \cite{fifty2021tag} that measures task affinities for task grouping, we employ affinities among multiple simultaneous FL tasks for splitting via four stages: 1) Each client measures affinities among FL tasks during \textit{all-in-one} training every $\rho$ batchs and averages them over $E$ local epoch; 2) The server obtains affinity scores by aggregating the affinities over $K$ participating clients; 3) The system computes the affinity scores of different combinations of subsets of FL tasks and select the best subset that achieve the highest affinity score; 4) The server splits the all-in-one model $\phi$ following the combination of tasks and continues training each split with its model initialized with parameters obtained from all-in-one training. Particularly, during training of all-in-one FL task $\alpha_0$, we measure the affinity of FL task $\alpha_i$ onto $\alpha_j$ at time step $t$ in client $k$ with the equation:
\begin{equation}
  \mathcal{S}^{k,t}_{\alpha_i \rightarrow \alpha_j} = 1 - \frac{\mathcal{L}_{\alpha_j}(\mathcal{X}^{k,t}, \theta_{s,\alpha_i}^{k,t+1}, \theta_{\alpha_j}^{k,t})}{\mathcal{L}_{\alpha_j}(\mathcal{X}^{k,t}, \theta_s^{k,t}, \theta_{\alpha_j}^{k,t})},
  \label{eq:single-affinity}
\end{equation}
where $\mathcal{L}_{\alpha_j}$ is the loss function of $\alpha_j$, $\mathcal{X}^{k,t}$ is a batch of training data, and $\theta_s^{k,t}$ and $\theta^{k,t+1}_{s,\alpha_i}$ are the shared model parameters \emph{before} and \emph{after} updated by $\alpha_i$, respectively. A positive value of $\mathcal{S}^{k,t}_{\alpha_i \rightarrow \alpha_j}$ means that task $\alpha_i$ helps reduce the loss of $\alpha_j$; the higher value of $\mathcal{S}^{k,t}_{\alpha_i \rightarrow \alpha_j}$ suggests that these two tasks are better to train together. This equation measures the affinity of one time step of one client. We approximate affinity scores for each round by averaging the values over $T$ time steps in $E$ local epochs and $K$ selected clients:
  $\hat{\mathcal{S}}_{\alpha_i \rightarrow \alpha_j} = \frac{1}{K E T} \sum_{k=1}^K \sum_{e=1}^E \sum_{t=1}^T \mathcal{S}^{k,t}_{\alpha_i \rightarrow \alpha_j}$,
where $T$ is the total time steps determined by the frequency $\rho$ of calculating Equation \ref{eq:single-affinity}, e.g., $\rho = 5$ means measuring the affinity in each client in every five batches.

These affinity scores measure pair-wise affinities between FL tasks. We next use them to calculate total affinity scores of a split with $\sum_{i=1}^n \hat{\mathcal{S}}_{\alpha_i}$, where $\hat{\mathcal{S}}_{\alpha_i}$ is the averaged affinity score onto each FL task. For example, a grouping of two splits among five FL tasks is $\{\alpha_1, \alpha_2\} and \{\alpha_3, \alpha_4, \alpha_5\}$, where $\{,\}$ denotes a split. The affinity score onto $\alpha_1$ is $\hat{\mathcal{S}}_{\alpha_1}=\hat{\mathcal{S}}_{\alpha_2 \rightarrow \alpha_1}$ and the affinity score onto $\alpha_3$ is ${\hat{\mathcal{S}}_{\alpha_3}} = (\hat{\mathcal{S}}_{\alpha_4 \rightarrow \alpha_3} + \hat{\mathcal{S}}_{\alpha_5 \rightarrow \alpha_3})/2$. Consequently, we can find the set $I$ with $|I|$ elements for subsets of $\mathcal{A}$ that maximize $\sum_{i=1}^n \hat{\mathcal{S}}_{\alpha_i}$, where $|I|$ defines the number of elements.

\begin{algorithm}[t]
  \centering
  \caption{Our Proposed MAS}\label{algo:mufl}
  \begin{algorithmic}[1]
    \State \textbf{Input:} FL tasks $
    \mathcal{A} = \{\alpha_1,\dots,\alpha_n\}$, available clients $\mathcal{C}$, number of selected clients $K$, local epoch $E$, aggregation weight of client $k$ $p_k$, training rounds $R$, all-in-one training rounds $R_0$, the number of splits $x$, frequency of computing affinities $\rho$, batch size $B$
    \State \textbf{Output:} models $\mathcal{W} = \{\omega_1,\omega_2,\dots,\omega_n\}$
    \State
    \State \textbf{\underline{ServerExecution:}}
    \State Receive FL tasks $\mathcal{A}$ and merge models $\mathcal{W}$ into a multi-task model $\phi^0 = \{\theta_s\} \cup \{\theta_{\alpha_i}\vert \alpha_i \in \mathcal{A} \}$ \Comment{Merging}
    \State Initialize $\phi^0$
    \For{\textit{each round} $r = 0, 1, ..., R_0-1$}
       \State $\mathcal{C}^r \gets$ (Randomly select K clients from $\mathcal{C}$)
       \For{\textit{client} $k \in \mathcal{C}^r$ \textit{in parallel}}
           \State $\phi^{k,r}, \hat{\mathcal{S}}_{\alpha_i \rightarrow \alpha_j}^{k,r} \gets$ \underline{\textbf{ClientExecution}}($\phi^r$, $\mathcal{A}$, $\rho$)
       \EndFor
       \State $\phi^{r+1} \gets \sum\limits_{k \in \mathcal{C}^r} p_k \phi^{k,r}$
       \State $\hat{\mathcal{S}}^r_{\alpha_i \rightarrow \alpha_j} \gets \frac{1}{K} \sum\limits_{k \in \mathcal{C}^r} \hat{\mathcal{S}}^{k,r}_{\alpha_i \rightarrow \alpha_j}$
    \EndFor

    \State Compute self-affinity $\hat{\mathcal{S}}^r_{\alpha_i \rightarrow \alpha_i}$ using Eqn. \ref{eq:self-affinity}  
    \State Compute a disjoint partition set $I$ of FL tasks $\mathcal{A}$ for $x$ splits $\{\mathcal{A}_j | j \in I \}$ that maximizes $\hat{\mathcal{S}}^r_{\alpha_i}$ using affinity scores $\hat{\mathcal{S}}^r_{\alpha_i \rightarrow \alpha_j}$, $\forall \alpha_i$, $\alpha_j \in \mathcal{A}$  \Comment{Splitting}

    \For{\textit{each element $j \in I$}}   \Comment{Schedule to train}
      \State Initialize $\phi_j = \{\theta_s^j\} \cup \{\theta_{\alpha_i} | \alpha_i \in \mathcal{A}_j \}$ with parameters of $\phi$
      \For{\textit{each round} $r = 0, 1, ..., R - R_0-1$}
        \State $\mathcal{C}^r \gets$ (Random select K from $\mathcal{C}$)
        \For{\textit{client} $k \in \mathcal{C}^r$ \textit{in parallel}}
            \State $\phi_j^{k,r},$ \_ $\gets$ \underline{\textbf{ClientExecution}}($\phi_j^r$, $\mathcal{A}_j$, 0)
        \EndFor
        \State $\phi_j^{r+1} \gets \sum\limits_{k \in \mathcal{C}^r} p_k \phi_j^{k,r}$
      \EndFor
    \EndFor

    \State Reconstruct $\mathcal{W} = \{\omega_1, \omega_2, \dots, \omega_n\}$ from $\{\phi_j|j \in I\}$

    \State \textbf{Return}  $\mathcal{W}$
    \State

   \State\underline{\textbf{ClientExecution}} ($\phi$, $\mathcal{A}$, $\rho$):
   \State $T = \lfloor\frac{B}{\rho}\rfloor$ \textbf{if} $\rho \neq 0$ \textbf{else} $0$
   \For{\textit{local epoch} $e = 1, ..., E$}
      \State Update model parameters $\phi$ w.r.t FL tasks $\mathcal{A}$
       \For{each time-step $t = 1, ..., T$ (every $\rho$ batches)}
          \State Compute $\mathcal{S}^t_{\alpha_i \rightarrow \alpha_j}$ using Eqn. \ref{eq:single-affinity}, $\forall \alpha_i$, $\alpha_j \in \mathcal{A}$
       \EndFor
   \EndFor
   \State $\hat{\mathcal{S}}_{\alpha_i \rightarrow \alpha_j} = \frac{1}{ET} \sum\limits_{e=1}^{E} \sum\limits_{t=1}^{T} \mathcal{S}^{t}_{\alpha_i \rightarrow \alpha_j}$, $\forall \alpha_i$, $\alpha_j \in \mathcal{A}$

   \State \textbf{Return} $\phi$, $\hat{\mathcal{S}}_{\alpha_i \rightarrow \alpha_j}$
  \end{algorithmic}
\end{algorithm}

It is important to note the differences between our method and TAG \cite{fifty2021tag}. Firstly, TAG focuses on inference efficiency, thus it allows overlapping task grouping that could train one task multiple times. In contrast, our focus is fundamentally different: we focus on training efficiency and consider only non-overlapping splitting of FL tasks. Secondly, TAG is computation-intensive for higher numbers of splits, e.g., it fails to produce results of five splits of nine tasks in a week, whereas we only need seconds of computation. Thirdly, TAG rules out the possibility that a split contains only one task. The calculated value of $\hat{\mathcal{S}}{\alpha_i \rightarrow \alpha_i}$ from Equation \ref{eq:single-affinity} is larger than the values of $\hat{\mathcal{S}}{\alpha_i \rightarrow \alpha_j}$, where $i \neq j$. As a result, one task $\alpha_i \in \mathcal{A}$ consistently receives the highest score during splitting and is always assigned as a group. TAG sets $\hat{\mathcal{S}}_{\alpha_i \rightarrow \alpha_i} = 1\mathrm{e}^{-6}$, resulting in no group contains only a single task because the scores of other combinations are larger than $1e^{-6}$. To overcome these issues, we propose a new method to calculate the value as follows:
\begin{equation}
  \hat{\mathcal{S}}_{\alpha_i \rightarrow \alpha_i} = \sum_{j \in \mathcal{N}\backslash\{i\}} \frac{(\hat{\mathcal{S}}_{\alpha_i \rightarrow \alpha_j} + \hat{\mathcal{S}}_{\alpha_j \rightarrow \alpha_i})}{2n-2},
  \label{eq:self-affinity}
\end{equation}
where $\mathcal{N}=\{1,2,\dots,n\}$. The intuition of this equation is to measure the normalized affinity of task $\alpha_i$ to other tasks and other tasks to $\alpha_i$, thus, it is dubbed self-affinity. Equation \ref{eq:self-affinity} overrides the values in Equation \ref{eq:single-affinity} in affinity score calculation. Fourthly, we focus on training multiple simultaneous FL tasks, thus, we further aggregate affinity scores over $K$ selected clients. Finnaly, TAG trains each set $\mathcal{A}_j$ from scratch, whereas we initialize their models with the parameters obtained from all-in-one training. Table \ref{tab:comparison-optimal-worst} shows that this change significantly boosts the performance.

\section{Experiments}
\label{sec:experiments}

We evaluate the performance and resource usage of MAS and study the following questions: 1) How effective are the splits from MAS? 2) When to split the all-in-one FL task? 3) How much MAS can outperform the baseline of training each client independently? 4) What are the impacts of local epoch and the number of selected clients?

\subsection{Experiment Setup}

\begin{figure}[h]
  \begin{center}
  \includegraphics[width=\linewidth]{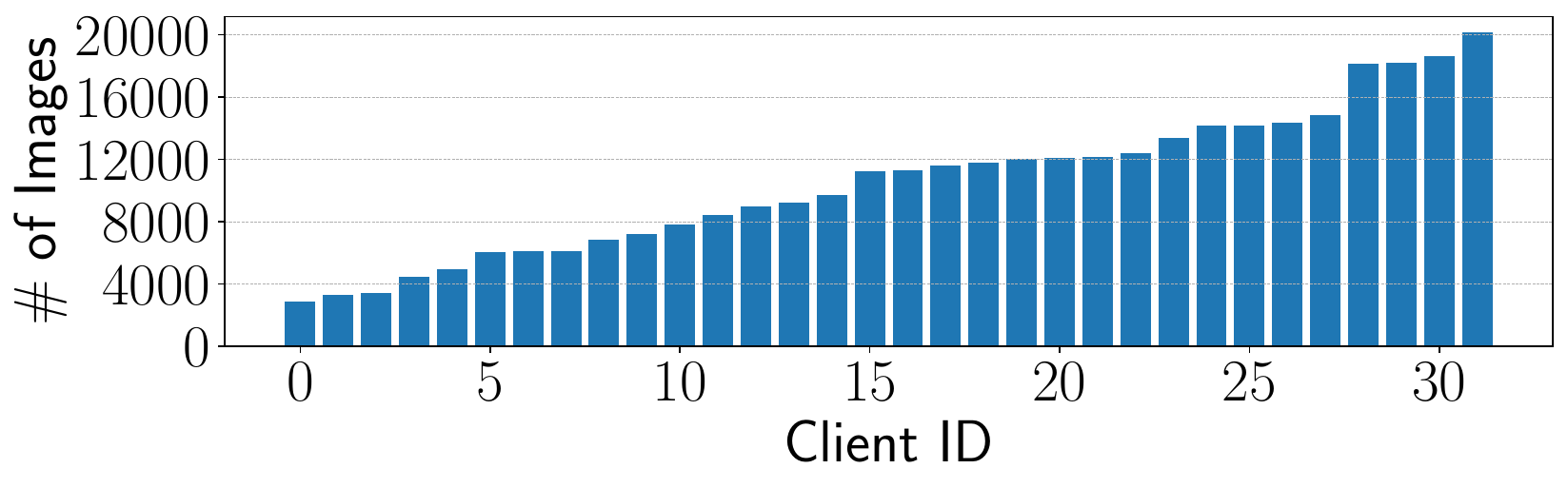}
     \caption{The data amount distribution of 32 FL clients. It simulates the challenging statistical heterogeneity.}
  \label{fig:client-stats}
  \end{center}
\end{figure}

\textbf{Dataset and Federated Simulation.} \, We construct our experiments using Taskonomy dataset \cite{zamir2018taskonomy}, which is a large and challenging computer vision dataset of indoor scenes of buildings. We run experiments with $N = 32$ clients, where each client contains a dataset of one building to simulate the statistical heterogeneity. Figure \ref{fig:client-stats} shows the data amount distribution over 32 clients; some clients have only 4,000 images, whereas some clients have over 16,000 images. We design three sets of FL tasks to evaluate the robustness of MAS under different combintations and different numbers of CV tasks. These three sets are \texttt{sdnkt}, \texttt{erckt}, and \texttt{sdnkterca}; each character represents an FL task\footnote{The meaning of each character in \texttt{sdnkterca} are as follows; \texttt{s}: semantic segmentation, \texttt{d}: depth estimation, \texttt{n}: normals, \texttt{k}: keypoint, \texttt{t}: edge texture, \texttt{e}: edge occlusion, \texttt{r}: reshaping, \texttt{c}: principle curvature, \texttt{a}: auto-encoder.}. The \texttt{sdnkterca} set is especially challenging with 9 FL tasks.

\textbf{Implementation Details.} \, We implement our proposed MAS by PyTorch \cite{paszke2017pytorch} and EasyFL \cite{zhuang2022easyfl}. We simulate the FL training on a cluster of NVIDIA Tesla V100 GPUs, where each node in the cluster contains 8 GPUs. In each round, each selected client is allocated to a GPU to conduct training; these clients communicate via the NCCL backend. Besides, we employ FedAvg \cite{fedavg} for the server aggregation. By default, we randomly select $K = 4$ clients to train for $E = 1$ local epochs in each round and train for $R = 100$ rounds. The batch size is $B = 64$ for \texttt{sdnkt} and \texttt{erckt} and $B = 32$ for \texttt{sdnkterca}. We use the modified Xception Network \cite{chollet2017xception} as the encoder for FL tasks of \texttt{sdnkt} and \texttt{erckt}, and \textit{half size} of the network (half amount of parameters) for FL tasks of \texttt{sdnkterca}. The decoders contain four deconvolution layers and four convolution layers. The optimizer is stochastic gradient descent (SGD), with momentum of 0.9 and weight decay $1e^{-4}$. The learning rate is initiated as $\eta = 0.1$ and is updated with polynomial learning rate decay $(1 - \frac{r}{R})^{0.9}$ in each round, where $r$ is the number of trained rounds. We measure the statistical performance of a FL task set using the sum of test losses and measure the training time and energy consumption \cite{anthony2020carbontracker}. Most of the experimental results are from three independent runs. More experimental details are provided in the supplementary.

\begin{figure}[t!]
  \centering
  \begin{subfigure}[t]{0.49\textwidth}
      \includegraphics[width=\textwidth]{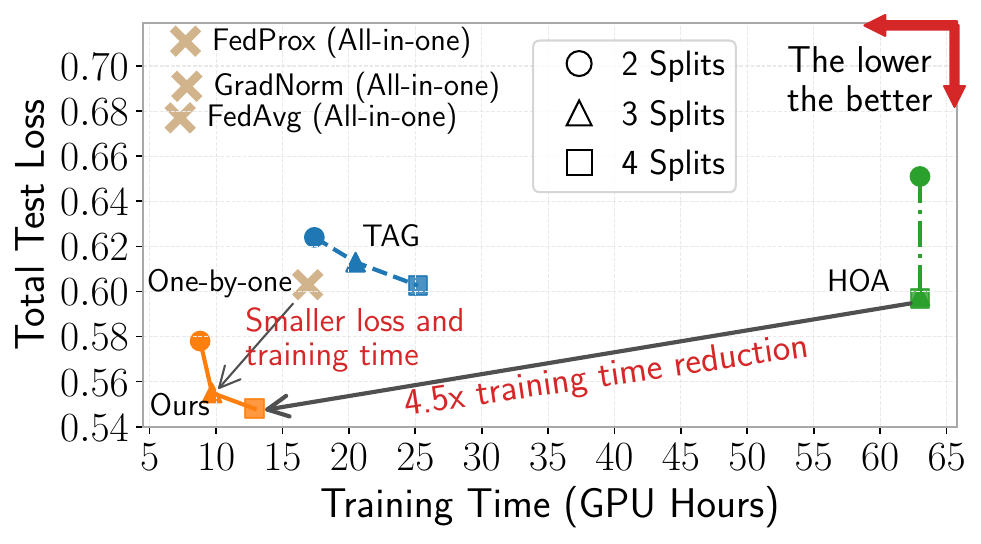}
      \label{fig:sdnkt-resource}
  \end{subfigure}
  \hfill
  \begin{subfigure}[t]{0.49\textwidth}
      \begin{tabular}{c|c|c|c}

Method & Test Loss & Training Time & Energy \\
& & (GPU $\times$ hours) & (Kwh)  \\\hline
One-by-one  & 0.603 \footnotesize{± 0.030} & 16.9 \footnotesize{± 0.5} & 8.4 \footnotesize{± 0.1} \\\hline
FedAvg \footnote{\label{tf}All-in-one methods} & 0.677 \footnotesize{± 0.018} & \textbf{7.3} \footnotesize{± 0.3} & \textbf{3.7} \footnotesize{± 0.1} \\
FedProx \footref{tf} & 0.711 \footnotesize{± 0.070} & 7.7\footnotesize{± 0.5} & 4.4 \footnotesize{± 0.7} \\
GradNorm \footref{tf} & 0.691 \footnotesize{± 0.013} & 7.8 \footnotesize{± 0.6} & 4.1 \footnotesize{± 0.4} \\\hline
HOA-2 & 0.651 \footnotesize{± 0.029} & 63.0 \footnotesize{± 0.9} & 31.0 \footnotesize{± 0.5} \\
HOA-3 & 0.598 \footnotesize{± 0.029} & 63.0 \footnotesize{± 0.9} & 31.0 \footnotesize{± 0.5} \\
HOA-4 & 0.597 \footnotesize{± 0.015} & 63.0 \footnotesize{± 0.9} & 31.0 \footnotesize{± 0.5} \\\hline
TAG-2 & 0.624 \footnotesize{± 0.015} & 17.4 \footnotesize{± 0.5} & 9.8 \footnotesize{± 0.3} \\
TAG-3 & 0.613 \footnotesize{± 0.032} & 20.5 \footnotesize{± 0.7} & 11.3 \footnotesize{± 0.2} \\
TAG-4 & 0.603 \footnotesize{± 0.027} & 25.2 \footnotesize{± 0.8} & 13.7 \footnotesize{± 0.3} \\\hline
\textbf{MAS-2} & \textbf{0.578} \footnotesize{± 0.015} & 8.8 \footnotesize{± 0.5} & 4.9 \footnotesize{± 0.3} \\
\textbf{MAS-3} & \textbf{0.555} \footnotesize{± 0.015} & 9.7 \footnotesize{± 0.5} & 5.4 \footnotesize{± 0.3} \\
\textbf{MAS-4} & \textbf{0.548} \footnotesize{± 0.001} & 12.9 \footnotesize{± 0.6} & 6.7 \footnotesize{± 0.3} \\
\end{tabular}
     \label{fig:sdnkt-table}
 \end{subfigure}
 \caption{Comparison of test loss, training time, and energy consumption on a set of five FL tasks \texttt{sdnkt}, where each character represents an FL task. The figure visualizes the table on test loss and training time. Our proposed MAS achieves the best performance with only slight increases in training time than all-in-one methods, but it requires significant less training time than the other methods.}
 \label{fig:performance-vs-resource}
\end{figure}

\begin{figure}[t!]
  \centering
  \begin{subfigure}[t]{0.23\textwidth}
      \includegraphics[width=\textwidth]{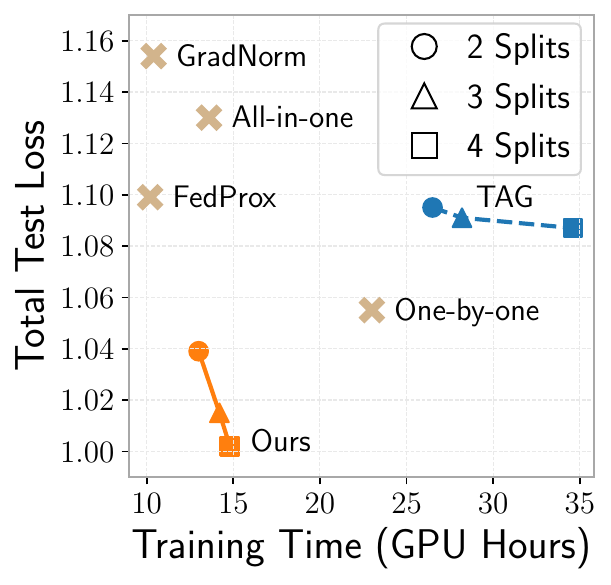}
      \caption{Five tasks: \texttt{erckt}}
      \label{fig:erckt-resource}
  \end{subfigure}
  \hfill
  \begin{subfigure}[t]{0.23\textwidth}
     \includegraphics[width=\textwidth]{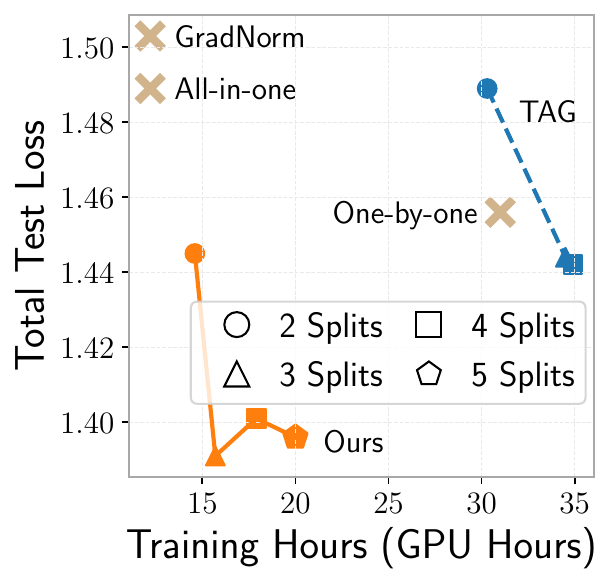}
     \caption{Nine tasks: \texttt{sdnkterca}}
     \label{fig:sdnkterca-resource}
 \end{subfigure}
 \caption{Comparison of test loss and training time on two FL task sets: (a) \texttt{erckt} and (b) \texttt{sdnkterca}. Our method achieves the best performance with only a slight increase in training time than all-in-one methods and much less training time than the other methods.}
 \label{fig:performance-vs-resource-2}
\end{figure}
\subsection{Performance Evaluation}
 
\begingroup
\setlength{\tabcolsep}{0.43em}
\begin{table*}[t]\centering
  \begin{tabular}{ccccccccc}\hline
    Task &\multirow{2}{*}{Splits} &\multirow{2}{*}{Ours} &\multicolumn{2}{c}{Train from Scratch} & &\multicolumn{2}{c}{Train from Initialization} \\\cline{4-5}\cline{7-8}
    Set & & &Optimal &Worst & &Optimal &Worst \\
    \hline
    \multirow{2}{*}{\texttt{sdnkt}} &2 &\textbf{0.578 $\pm$ 0.015} &0.622 $\pm$ 0.007 &0.685 $\pm$ 0.010 & &0.595 $\pm$ 0.008 &0.595 $\pm$ 0.004 \\
    &3 &\textbf{0.555 $\pm$ 0.008} &0.585 $\pm$ 0.026 &0.674 $\pm$ 0.022 & &0.560 $\pm$ 0.006 &0.578 $\pm$ 0.006 \\
    \hline
    \multirow{2}{*}{\texttt{erckt}} &2 &\textbf{1.039 $\pm$ 0.024} &1.070 $\pm$ 0.013 &1.312 $\pm$ 0.065 & &1.048 $\pm$ 0.024 &1.068 $\pm$ 0.037 \\
    &3 &\textbf{1.015 $\pm$ 0.018} &1.058 $\pm$ 0.029 &1.243 $\pm$ 0.099 & &1.020 $\pm$ 0.012 &1.052 $\pm$ 0.026 \\
    \hline
  \end{tabular}
  \caption{Performance (total test loss) comparison of our proposed MAS with the optimal and worst splits. Our method achieves the best performance, indicating the effectiveness of the splits obtained from MAS.}
  \label{tab:comparison-optimal-worst}  
\end{table*}
\endgroup

We conduct the performance comparison, in terms of total test loss of all tasks, training time, and energy consumption, among the following methods: 1) one-by-one training of FL tasks; 2) all-in-one training of FL tasks using FedAvg; 3) all-in-one training with
multi-task optimization (GradNorm \cite{chen2018gradnorm}) and federated optimization (FedProx \cite{fedprox}); 4) HOA \cite{standley2020which} method that groups FL tasks by estimating higher-order of groupings from pair-wise tasks performance and then trains each group from scratch; 5) TAG \cite{fifty2021tag} method that groups FL tasks only based on task affinity and trains each group from scratch; 6) Our proposed MAS.

 Figure \ref{fig:performance-vs-resource} compares the performance of the above methods on a set of five simultaneous FL tasks \texttt{sdnkt}. The methods that achieve lower total test loss and lower energy consumption are better. At the one extreme, all-in-one methods consume the least training time and energy, but their test losses are the highest. Simply applying multi-task learning optimization (GradNorm \cite{chen2018gradnorm}) or federated optimization (FedProx \cite{fedprox}), can hardly improve performance. At the other extreme, HOA can achieve comparable test losses, but it demands long training time and high energy consumption (around $4-6\times$ of ours) to compute pair-wise tasks for higher-order estimation. Although the one-by-one method and TAG \cite{fifty2021tag} present a good balance between test loss and system metric, MAS is superior in both aspects; it achieves the best test loss with $\sim$2$\times$ reduction on training time and $\sim$40\% less energy consumption. We evaluate the performance of splitting the all-in-one FL task into 2, 3, and 4 splits in our method, denoted as MAS-2, MAS-3, and MAS-4, respectively. More splits lead to longer training time, but it could help further reduce test losses. 
 
 In addition, Figure \ref{fig:performance-vs-resource-2} further compares the performance of total test loss and training time on another five-task set \texttt{erckt} and a nine-task set \texttt{sdnkterca}. They generally achieve similar results as task set \texttt{sdnkt}. In set \texttt{sdnkterca}, we further find that splitting into more splits may not always improve the total test loss, though they are still better than other methods. The reason could be that it is easier to find tasks that have better synergies when training with more simultaneous tasks. HOA on \texttt{erckt} is at least 5.5$\times$ slower than our method. We do not report HOA for \texttt{sdnkterca} due to computation constraints; HOA computes at least 36 pairs of FL tasks ($\sim$720 GPU hours).

 \subsection{How Effective are the Splits From MAS?} 

 \begin{figure}[t]
  \centering
  \begin{subfigure}[t]{0.15\textwidth}
    \includegraphics[width=\textwidth]{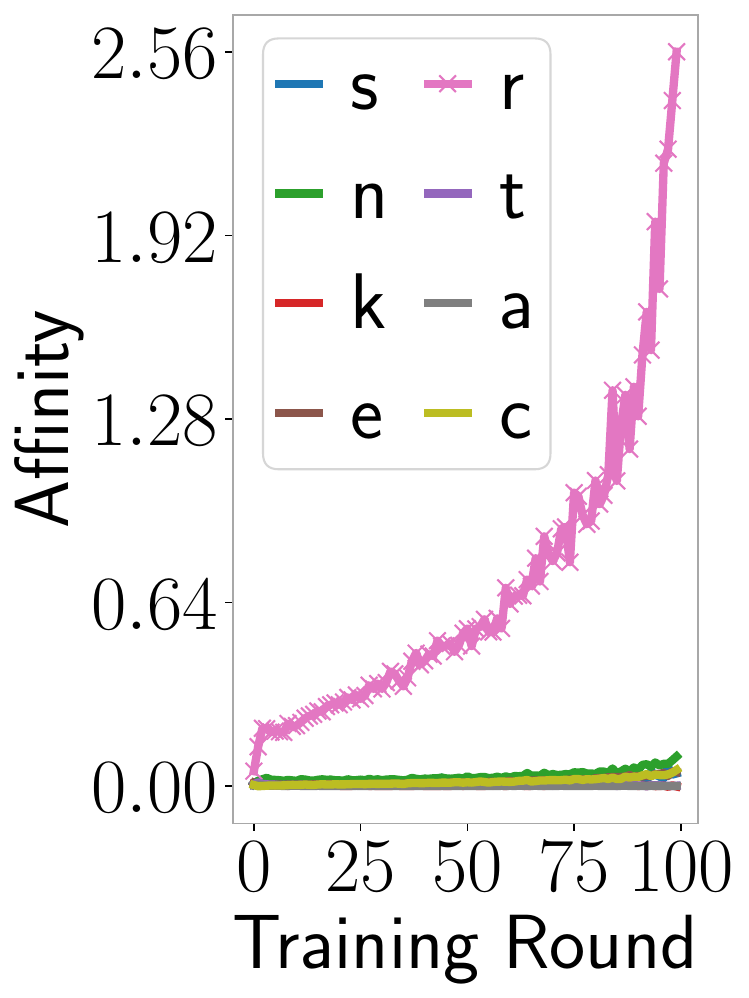}
    \caption{Affinities to \texttt{d}}
    \label{fig:affinity-to-d}
  \end{subfigure}
  \hfill
  \begin{subfigure}[t]{0.15\textwidth}
    \includegraphics[width=\textwidth]{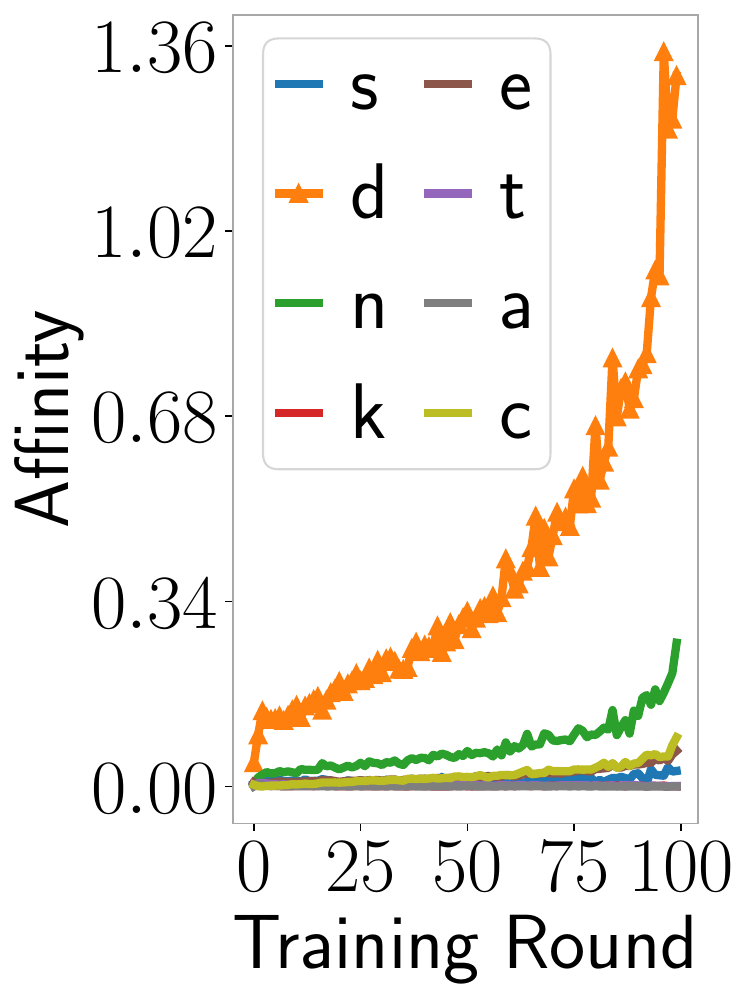}
    \caption{Affinities to \texttt{r}}
    \label{fig:affinity-to-r}
  \end{subfigure}
  \hfill
  \begin{subfigure}[t]{0.15\textwidth}
    \includegraphics[width=\textwidth]{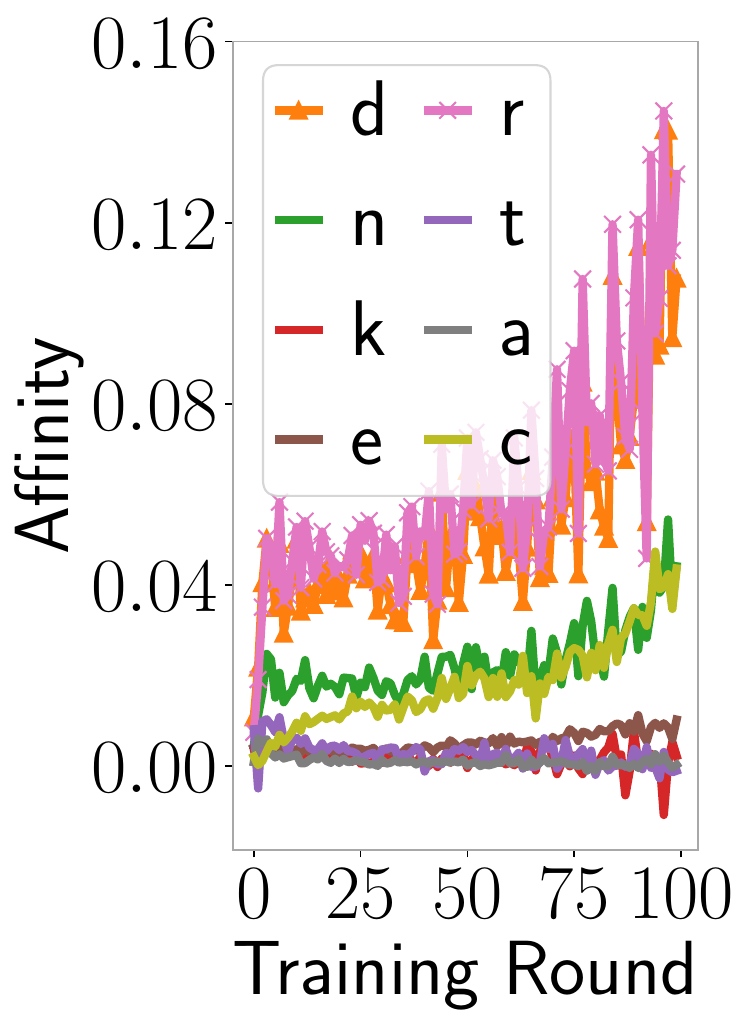}
    \caption{Affinities to \texttt{s}}
    \label{fig:affinity-to-s}
  \end{subfigure}
 \caption{Changes of affinity scores of one FL task to the other on task set \texttt{sdnkterca}. FL task \texttt{d} and \texttt{r} have high inter-task scores. The trends of affinities emerge at the early stage of training.}
 \label{fig:affinity-analysis}
\end{figure}
 
 We demonstrate the effectiveness of the splitting method in MAS by comparing it with the performance of splits with possible optimal and worst splits. The optimal and worst splits are obtained with two steps: 1) we measure the performance over all combinations of two splits and three splits of an FL task set by training them from scratch;\footnote{There are fifteen and twenty-five combinations of two and three splits, respectively, for a set of five tasks.} 2) we select the combination that yields the best performance as the optimal split and the worst performance as the worst split.
 
 Table \ref{tab:comparison-optimal-worst} compares the total test loss of MAS with the optimal and worst splits trained in two ways: 1) training each split from scratch; 2) training each split based on model parameters obtained from all-in-one training, which is adopted in our method but not in TAG \cite{fifty2021tag}. On the one hand, training from initialization outperforms training from scratch in all settings. It suggests that initializing each split with all-in-one training model parameters can significantly improve the performance. On the other hand, our splitting method achieves the best performance in all settings, even though training from initialization reduces the gaps of different splits (the optimal and worst splits). These results indicate the effectiveness of the splits obtained from MAS.

 \begin{figure}[t!]
    \centering
    \begin{subfigure}[t]{0.1475\textwidth}
      \includegraphics[width=\textwidth]{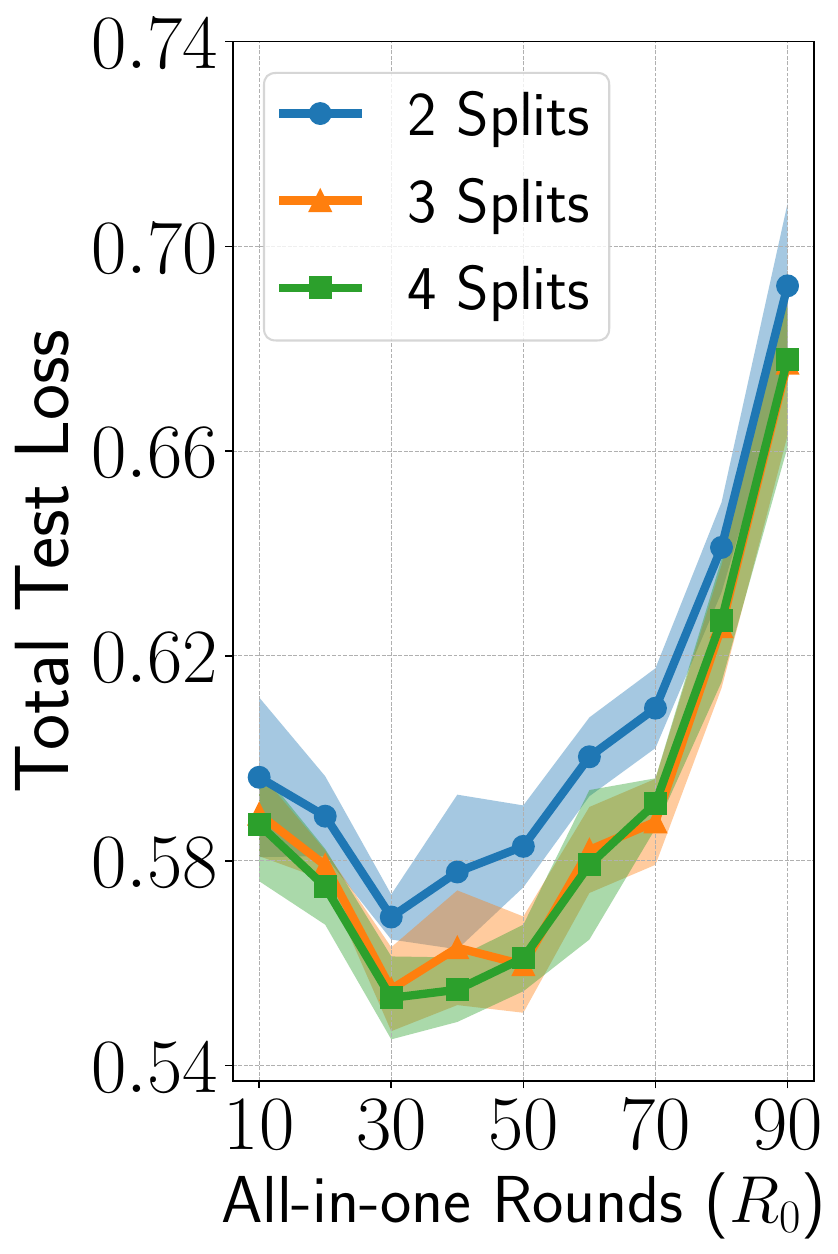}
      \caption{Tasks: \texttt{sdnkt}}
      \label{fig:sdnkt-rounds}
    \end{subfigure}
    \begin{subfigure}[t]{0.1475\textwidth}
      \includegraphics[width=\textwidth]{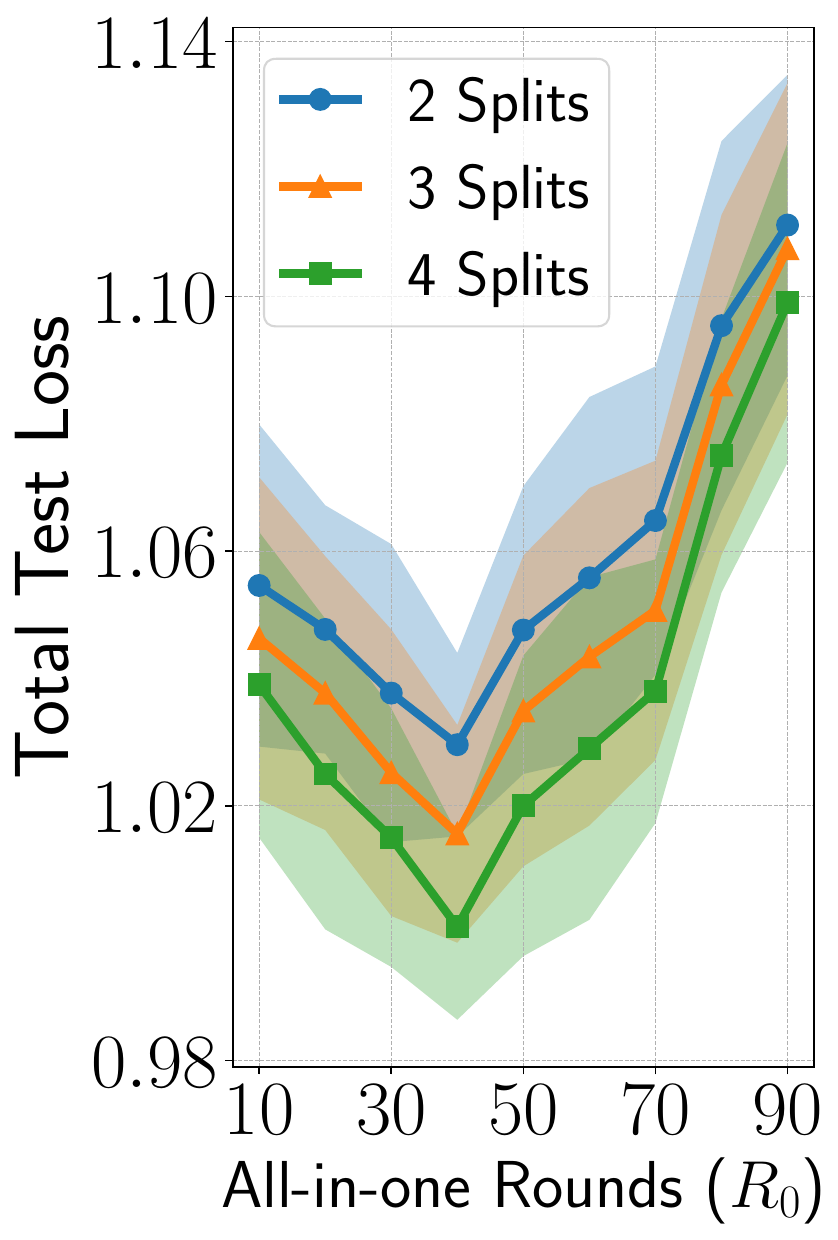}
      \caption{Tasks: \texttt{erckt}}
      \label{fig:erckt-rounds}
    \end{subfigure}
    \hfill
    \begin{subfigure}[t]{0.173\textwidth}
      \includegraphics[width=0.85\textwidth]{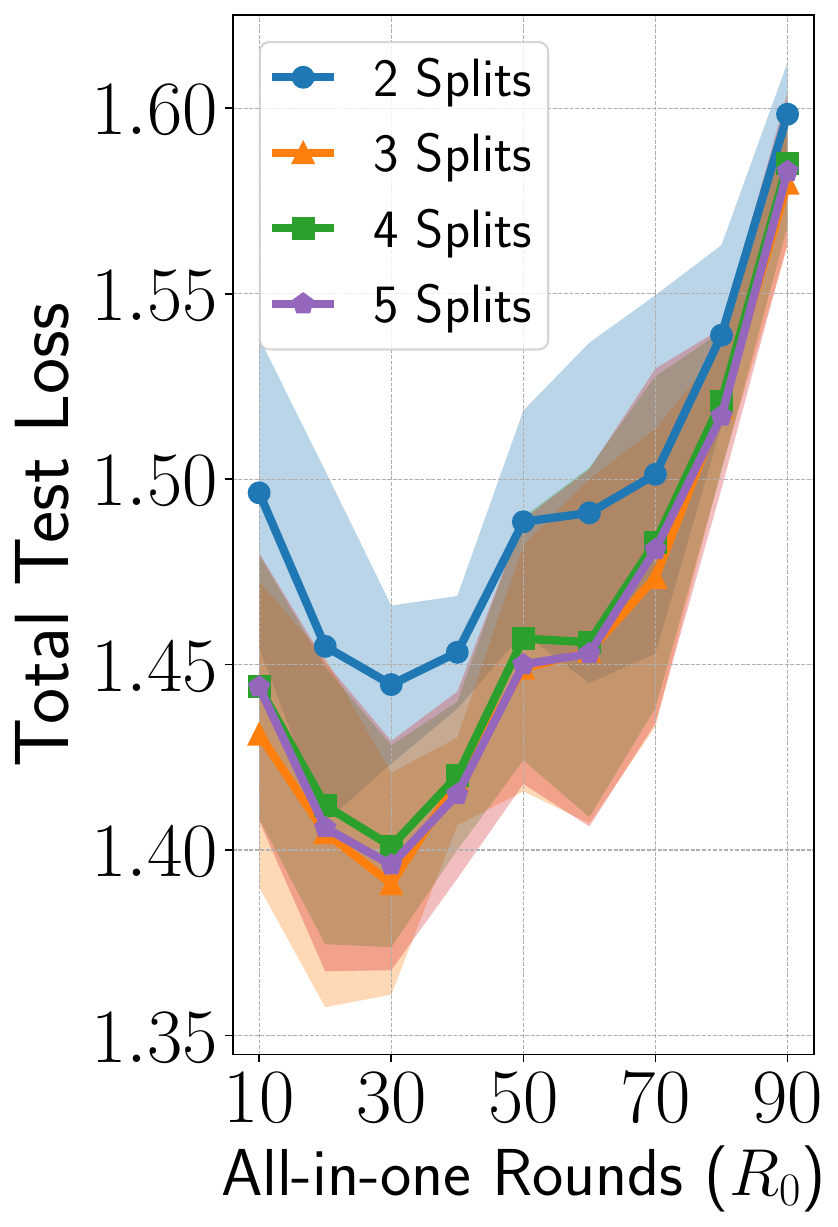}
      \caption{Tasks: \texttt{sdnkterca}}
      \label{fig:sdnkterca-rounds}
    \end{subfigure}
   \caption{Comparison of training all-in-one tasks for different $R_0$ rounds. Fixing the total training rounds $R = 100$, MAS achieves the best performance when $R_0 \in \{30, 40\}$.}
   \label{fig:all-in-one-rounds}
  \end{figure}

\subsection{When to Split the All-in-one FL Task?}

We further answer the question that how many $R_0$ rounds should we train the all-in-one FL task before splitting. It is determined by two factors: 1) the rounds needed to obtain affinity scores for a reasonable splitting; 2) the rounds that yield the best overall performance.
 
\textbf{Affinity Analysis.} \, We analyze changes in affinity scores over the course of training to show that MAS can use early-stage affinity scores for splitting. Figure \ref{fig:affinity-analysis} presents the affinity scores of different FL tasks to one FL task on task set \texttt{sdnkterca}. Figure \ref{fig:affinity-to-d} and \ref{fig:affinity-to-r} indicate that FL task \texttt{d} and FL task \texttt{r} have high inter-task affinity scores; they are split into the same group as a result. In contrast, both \texttt{d} and \texttt{r} have high-affinity score to FL task \texttt{s} in Figure \ref{fig:affinity-to-s}, but not vice versa. These trends emerge in the early stage of training, thus, we employ the affinity scores of the \emph{10-th} round for splitting by default; they are effective in achieving promising results as shown in Figure \ref{fig:performance-vs-resource} and \ref{fig:performance-vs-resource-2}. We provide more affinity scores of FL tasks in the supplementary.
 
\textbf{The Impact of $R_0$ Rounds.} \, Figure \ref{fig:all-in-one-rounds} compares the performance of training $R_0$ for 10 to 90 rounds before splitting. Fixing the total training round $R = 100$, we train each split of FL tasks for $R_1 = R - R_0$ rounds. The results indicate that MAS achieves the best performance when $R_0 = \{30, 40\}$ rounds. Training the all-in-one FL task for enough rounds helps utilize the benefits and synergies of training together, but training for too many rounds almost suppresses the benefits of considering differences among FL tasks. We suggest training $R_0$ for [30, 40] rounds to strike a good balance between these two extremes from the above empirical results and consider other mechanisms to determine $R_0$ in future works.

\subsection{Additional Analysis}

\begin{figure}[t!]
  \centering
   \begin{subfigure}[t]{0.18\textwidth}
    \includegraphics[width=\linewidth]{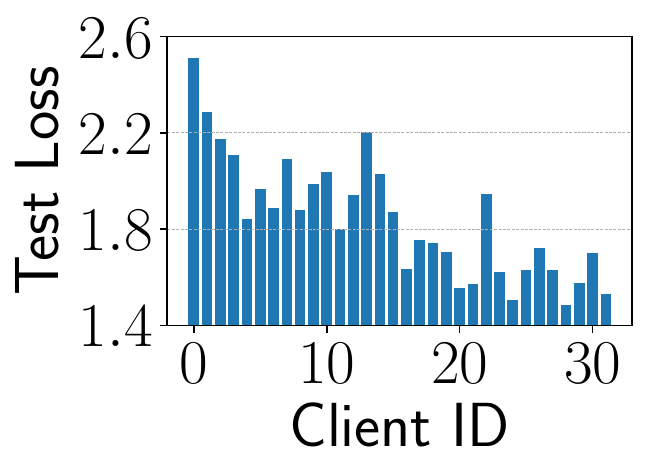}
    \caption{Test loss distribution of standalone training}
    \label{fig:loss-distribution}%
   \end{subfigure}
   \hfill
   \begin{subfigure}[t]{0.27\textwidth}
      \vspace{-2cm}
      \begin{tabular}{cccc}\hline
        Method &Total Test Loss \\\hline
        Standalone & 1.842 \\
        All-in-one & 0.677 \\
        MAS-2 & 0.578 \\\hline
        \end{tabular}
      \vspace{0.3cm}
      \caption{Test loss comparison}
      \label{fig:standalone-comparison}
   \end{subfigure}
  \caption{Performance of standalone training that conducts training using data in each client independently on task set \texttt{sdnkt}: (a) shows test loss distribution of 32 clients. (b) compares test losses of standalone training and FL methods.}
  \label{fig:standalone}
\end{figure}

We further analyze the performance of standalone training, the impact of local epoch $E$, and the impact of the number of selected clients $K$ in FL using all-in-one training. We report the results of FL task set \texttt{sdnkt} here and provide more results in the supplementary.

\textbf{Standalone Training.} \, Standalone training refers to training using data of each client independently. Figure \ref{fig:loss-distribution} shows the test loss distribution of 32 clients in experiments. The client ID corresponds to the dataset size distribution in Figure \ref{fig:client-stats}. These results suggest that clients with larger data sizes may not lead to higher performance. Figure \ref{fig:standalone-comparison} compares test losses of standalone training and FL methods. Either all-in-one or our MAS greatly outperforms standalone training. It suggests the significance of federated learning when data are not shareable among clients.

\begin{figure}[t]
  \centering
  \begin{subfigure}[t]{0.19\textwidth}
    \includegraphics[width=\textwidth]{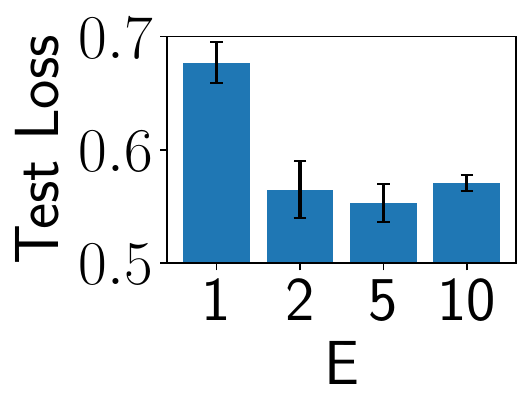}
    \caption{Impact of $E$}
    \label{fig:sdnkt-local-epoch}
  \end{subfigure}
  \begin{subfigure}[t]{0.182\textwidth}
    \includegraphics[width=\textwidth]{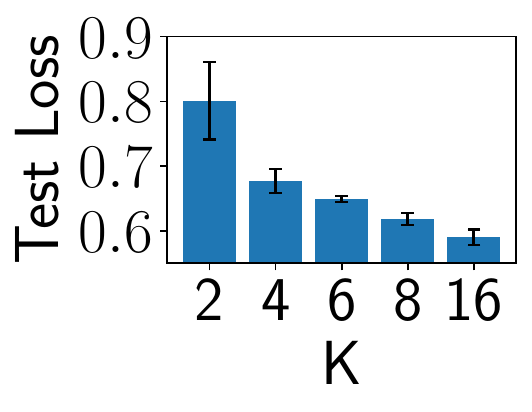}
    \caption{Impact of $K$}
    \label{fig:num-clients}
  \end{subfigure}
  \caption{Analysis of the impact of (a) local epoch $E$ and (b) the number of selected clients $K$ on task set \texttt{sdnkt}. Larger $E$ and $K$ could reduce losses with more computation, but the benefit decreases as computation increases.}
  \label{fig:ablation}
\end{figure}

\textbf{Impact of Local Epoch $E$.} \, Local epoch defines the number of epochs each client trains before uploading training updates to the server. Figure ~\ref{fig:sdnkt-local-epoch} compares test losses of local epochs $E = \{1, 2, 5, 10\}$. Larger $E$ could lead to better performance with higher computation (fixed training round $R=100$), but it is not effective when increasing $E=5$ to $E = 10$. It suggests the limitation of simply increasing computation with larger $E$ in improving performance. Note that MAS (Table \ref{tab:comparison-optimal-worst}) achieves better results than $E = 5$ with $\sim 5\times$ less computation.

\begin{table}[t]\centering
  \begin{tabular}{cccc}\hline
  Method &K &Total Test Loss \\\hline
  All-in-one & 8 & 0.618 \\
  MAS-2 & 8 & 0.512 \\
  \hline
  \end{tabular}
  \caption{Comparison of total test loss using $K = 8$ selected clients on FL task set \texttt{sdnkt}. MAS achieves even better performance on $K = 8$.}\label{tab:8-client-mufl}  
\end{table}

\textbf{Impact of The Number of Selected Clients $K$.} \, Figure \ref{fig:num-clients} compares test losses of the number of selected clients $K = \{2, 4, 6, 8, 16\}$ in each round. Increasing the number of selected clients improves the performance, but the effect becomes marginal as $K$ increases. Larger $K$ can also be considered as using more computation in each round. Similar to the results of the impact of $E$, simply increasing computation can only improve performance to a certain extent. It also shows the significance of MAS that increases performance with slightly more computation. 

The majority of experiments in this study are conducted with $K = 4$. We next analyze the impact of $K$ in MAS with results of two splits on task set \texttt{sdnkt} in Table \ref{tab:8-client-mufl}. The results indicate that MAS is also effective with $K = 8$, which outperforms $K = 4$ and all-in-one training.

\section{Conclusions}
\label{sec:conclusion}

In this work, we propose MAS, the first FL system to effectively coordinate and train multiple simultaneous FL tasks under resource constraints. In particular, we introduce task merging and task splitting to consider both synergies and differences among multiple FL tasks. Extensive empirical studies demonstrate that our method is effective in elevating performance and significantly reduce the training time and energy consumption by more than 40\%. We believe that it is important to study training multiple simultaneous FL tasks and apply it in many real-world applications. We hope this research will inspire the community to further work on algorithm and system optimizations of training multiple simultaneous FL tasks. Future work involves designing better scheduling mechanisms to coordinate these tasks. Client selection strategies can also be considered to optimize resource and training allocation.

\noindent\textbf{Acknowledgements} This study is supported by RIE2020 Industry Alignment Fund-Industry Collaboration Projects (IAF-ICP) Funding Initiative, as well as cash and in-kind contribution from the industry partner(s) and Sony AI.

{\small
\bibliographystyle{ieee_fullname}
\bibliography{egbib}
}

\end{document}